\documentclass{article}

     \usepackage[preprint]{neurips_2019}

\usepackage[utf8]{inputenc} 
\usepackage[T1]{fontenc}    
\usepackage{hyperref}       
\usepackage{url}            
\usepackage{booktabs}       
\usepackage{amsfonts}       
\usepackage{nicefrac}       
\usepackage{microtype}      
\usepackage{graphicx}
\usepackage{times}
\usepackage{epsfig}
\usepackage{graphicx}
\usepackage{amsmath}
\usepackage{amssymb}
\usepackage{lineno}
\usepackage{color}
\usepackage{subfigure}
\usepackage{multirow}
\usepackage{algorithm}
\usepackage{algorithmic}
\usepackage{epsfig}
\usepackage{caption}
\usepackage{bm}
\usepackage{url}
\usepackage{amsmath}

\title{Invasiveness Prediction of Pulmonary Adenocarcinomas Using Deep Feature Fusion Networks}

\author{
  Xiang Li \thanks{Equal contribution}\\
 Sun Yat-sen University \\
  \vspace{-0.1cm}
  \texttt{\scriptsize{lixiang651@gmail.com}}\\
   \And
   Jiechao Ma\footnotemark[1]\\
   Sun Yat-sen University \\
   \vspace{-0.1cm}
   \texttt{\scriptsize{majch7@mail2.sysu.edu.cn}} \\
   \And
   Hongwei Li \\
  Technical University of Munich \\
  \vspace{-0.1cm}
   \texttt{\scriptsize{hongwei.li@tum.de}}\\
}

\begin{document}

\maketitle

\begin{abstract}
  Early diagnosis of pathological invasiveness of pulmonary adenocarcinomas using computed tomography (CT) imaging would alter the course of treatment of adenocarcinomas and subsequently improve the prognosis. Most of the existing systems use either conventional radiomics features or deep-learning features alone to predict the invasiveness. In this study, we explore the fusion of the two kinds of features and claim that radiomics features can be complementary to deep-learning features. An effective deep feature fusion network is proposed to exploit the complementarity between the two kinds of features, which improves the invasiveness prediction results. We collected a private dataset that contains lung CT scans of 676 patients categorized into four invasiveness types from a collaborating hospital. Evaluations on this dataset demonstrate the effectiveness of our proposal.  
 
\end{abstract}

\section{Introduction}

Pulmonary adenocarcinoma is the most common histopathological subtype of lung cancer. Early identifying pathological invasiveness of pulmonary adenocarcinomas by computed tomography (CT) would be clinically important and could guide clinical decision making~\cite{maeshima2010histological,tsutani2013prognostic,yanagawa2017radiological,han2018ct}. According to the degrees of invasiveness, adenocarcinomas are classified as
atypical adenomatous hyperplasia (AAH), adenocarcinomas in situ (AIS), minimally invasive adenocarcinoma (MIA), and invasive adenocarcinoma (IA)~\cite{travis2011international}. Due to the noise and artifacts in CT imaging, it is challenging for radiologists to differentiate the degrees of invasiveness. The design of a reliable invasiveness prediction system is increasingly needed in clinical practice.

Existing works can be mainly categorised into two groups: methods that extract conventional radiomics features and identify the invasiveness types by using a statistical classifier such as support vector machine or random forest~\cite{xue2018use,mei2018predicting,zhao2019development}; methods that use convolutional neural network(CNN) which learn abstract features to predict invasiveness~\cite{zhao20183dcnn,yanagawa2019applicationcnn}. The radiomics features contain hand-crafted, low-level information while deep-learning features are data-driven high-level representation. 

In this paper, we investigate the combination and complementarity of radiomics features and deep-learning features. 
Specifically, we propose an effective deep feature fusion network which merges the radiomics and deep-learning features and is further optimized in a data-driven manner. Our model can incorporate the strength of these two different features to make an accurate prediction.

To the best of our knowledge, this is the first attempt to construct an end-to-end deep learning framework that combines radiomics features and deep-learning features to address the invasiveness prediction problem. Extensive experimental results on our private dataset demonstrate the effectiveness of the proposed framework and clearly indicate that complementarity modeling between different feature representations is valuable for identifying invasiveness.

\section{Methodology}

\begin{figure}[t]\vspace{-0cm}
	\begin{center}
		{
			\includegraphics[width=0.9\textwidth]{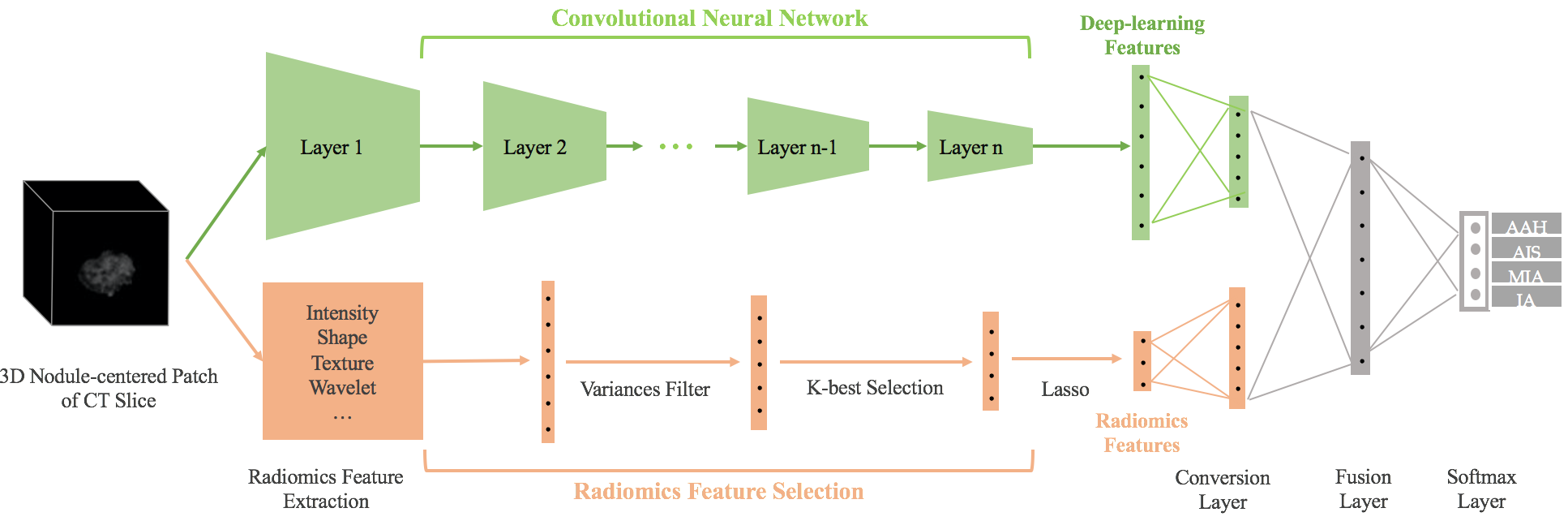}
		}
	\end{center}
	\vspace{-0cm}
	\caption{The architecture of our invasiveness prediction framework. The radiomics features and deep-learning features are aggregated in feature level and optimized in an end-to-end manner.}
	\label{framework} \vspace{-0.2cm}
\end{figure}

In this section, we present the proposed invasiveness prediction framework which is built based on an effective feature fusion network. The network
architecture is shown in Fig.~\ref{framework}. Our model consists of two streams,  
the first part deals with the convolutional neural network given 3D nodule-centered patches of lung CT slices as the input to compute deep-learning features; 
the second part extracts and selects radiomics features given the same patches. These two features are followed by a new conversion layer, then concatenated together to produce an enhanced representation in a new fusion layer. We claim that the radiomics part regularizes and serves as deep supervision during the learning process. Finally, softmax function is applied in the last layer to make label predictions, and the cross-entropy loss is minimized for training the network.

\noindent\textbf{Deep-learning features.} The ResNet50~\cite{he2016deep} architecture is used for extracting deep-learning features. Specifically, in order to capture spatial context information in neighboring CT slices, we utilize 3D convolution with the same padding instead of 2D convolution layer. The output of the average pooling layer is a 2048-dimensional vector, which we regarded as the deep-learning features.

\noindent\textbf{Radiomics features.} We firstly extract around 4000-dimensional radiomics features from the CT slices, which can be categorized into four types such as intensity, shape, texture and wavelet~\cite{van2017computational}. 
Next, we remove the radiomics characteristics with low variance (lower than 0.8) on the training set. Finally, the remaining radiomics features are processed by K-best method and Lasso model which select around 100-dimensional features. 

\noindent\textbf{Feature fusion.} The fusion layer utilizes full connection to provide self-adaptation on the combination of radiomics features ($\mathbf{RF}$) and deep-learning features ($\mathbf{DF}$). 
Before the fusion action, both features are followed by a new conversion layer, namely a 512D-output full connection layer, which bridges the dimensional difference between the two kinds of features, and improves the convergence of our feature fusion network.
Specifically, the input of fusion layer is $\mathbf{x} = \mathbf{[RF, DF]}$, then the output of this layer is computed by $f_{Fusion}(\mathbf{x}) = h(\mathbf{W}_{Fusion}\mathbf{x}+\mathbf{b}_{Fusion})$, 
where $h(\mathbf{\cdot})$ denotes the ReLU activation function. By this means, our model can jointly map these two different features to an embedding feature space and make better use of the individual feature strength.

\section{Experiments}

\begin{figure}[htb]
	\begin{center}
		\subfigure[AAH]{
			\includegraphics[width=0.22\textwidth]{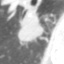}
		}
		\subfigure[AIS]{
			\includegraphics[width=0.22\textwidth]{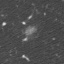}
		}
		\subfigure[MIA]{
			\includegraphics[width=0.22\textwidth]{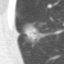}
		}
		\subfigure[IA]{
			\includegraphics[width=0.22\textwidth]{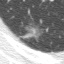}
		}
	\end{center}
	\vspace{-0cm}
	\caption{The examples of our dataset.}
	\label{dataset} \vspace{-0.4cm}
\end{figure}

\noindent\textbf{Dataset.} To evaluate the performance of our framework, we collected lung CT scans of 676 patients from a collaborating hospital. Each CT scan was manually segmented to obtain a 3D nodule-centered patch. The invasiveness of these nodules were pathology proved and graded as AAH (158 patients), AIS (136 patients), MIA (53 patients), and IA (329 patients). We randomly selected 80\% of patients as training set and 20\% of patients as testing set. Examples are shown in Fig.~\ref{dataset}.

\noindent\textbf{Training details and settings.} MXNet platform~\cite{chen2015mxnet} was applied to construct the proposed deep architecture. The model parameters were initialized by a pre-trained ResNet50 and then were fine-tuned around 20 epochs using the early stopping criterion. The stochastic gradient descent (SGD) optimizer was used with learning rate of 0.001. 

\noindent\textbf{Results and analysis.} The proposed model was evaluated on our private dataset by comparing with three different representative methods, which are marked as `RF+SVM', `CNN' and `RF+SVM+CNN', respectively. The results are shown in Table~\ref{results}. 
Specifically, 
`RF+SVM' denotes using radiomics features alone and a support vector machine classifier. 
`CNN' indicates using deep convolutional neural network (ResNet50) alone. 
`RF+SVM+CNN' represents the class probability combination of `RF+SVM' and `CNN'.

\begin{table}[htb]\vspace{-0.2cm}
	\centering
	\caption{The overall classification accuracy (\%) of different methods, which is calculated by comparing the prediction and ground-true label of each patient in testing set.}
	\setlength{\tabcolsep}{18pt}
	\begin{tabular}{c|c|c|c|c}\hline 
		Methods   &RF+SVM  &CNN  &RF+SVM+CNN & Ours  \\\hline
		Accuracy  &82.78  &86.09  &87.42 &\textbf{88.74}  \\
		\hline
	\end{tabular}
	\label{results}\vspace{-0.2cm}
\end{table}

\begin{figure}[htb]
	\begin{center}
		\subfigure[RF+SVM]{
			\includegraphics[width=0.22\textwidth]{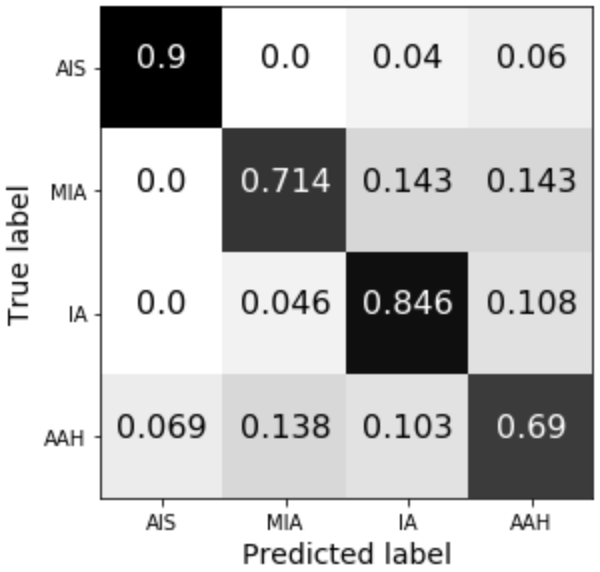}
		}
		\subfigure[CNN]{
			\includegraphics[width=0.22\textwidth]{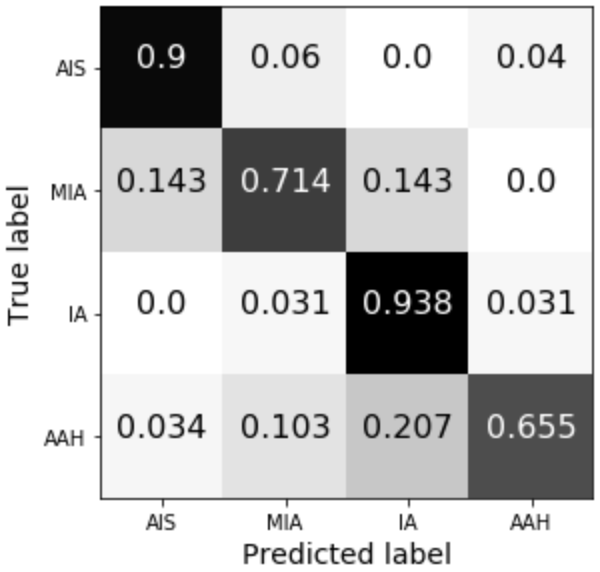}
		}
		\subfigure[RF+SVM+CNN]{
			\includegraphics[width=0.22\textwidth]{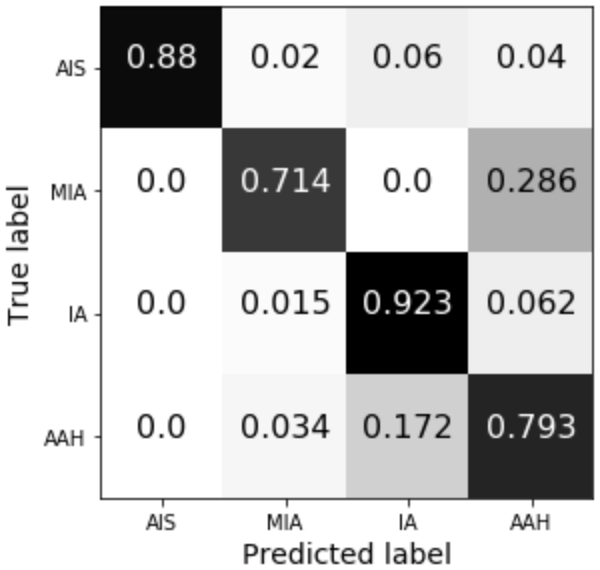}
		}
		\subfigure[Ours]{
			\includegraphics[width=0.26\textwidth]{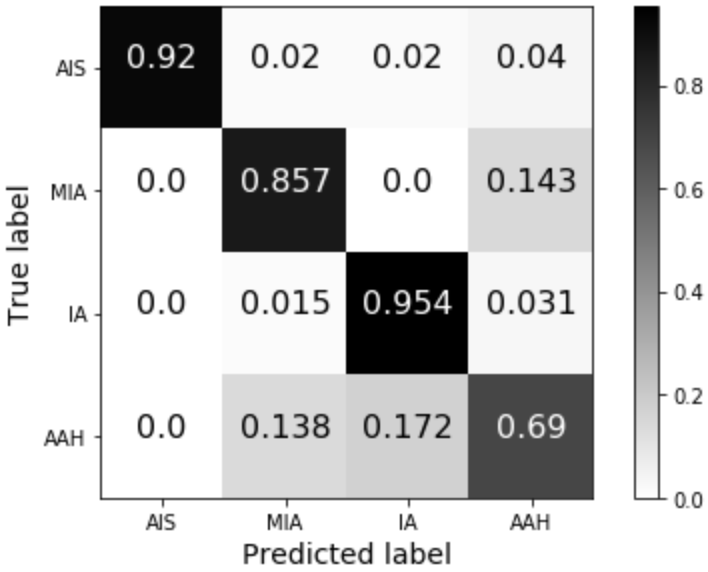}
		}
	\end{center}
	\vspace{-0.4cm}
	\caption{The confusion matrix of different methods.}
	\label{cm} \vspace{-0.2cm}
\end{figure}

As shown in Table~\ref{results}, the deeply fused feature not only outperforms radiomics features and deep-learning features alone, but also show its superiority over simple fusion strategy. Note that, `RF+SVM+CNN' performs the second-best and outperforms both `RF+SVM' and `CNN', which provides supports on our assumption that radiomics features and CNN features are complementary. Also, our model outperforms `RF+SVM+CNN' , due to the features are fused in the deep feature level and are further optimized, while in `RF+SVM+CNN', the predicted probabilities of CNN features are simply combined with the one of radiomics features, which may not be optimal.

We also visualized the confusion matrix results of different methods and show them in Fig.~\ref{cm}. We observed that the proposed fusion features outperformed or not worse than individual features in all classes while the simple combination strategy decreases the performance on the class AIS or falls to the middle performance on class IA, 
suggesting that the feature fusion network could effectively take advantage of different feature representations to improve the discrimination power.

\section{Conclusion}

In this work, we proposed an effective framework for invasiveness prediction of pulmonary adenocarcinoma, which jointly utilizes both the radiomics features and deep-learning features by training a feature fusion network.
This model uses radiomics features to regularize the convolutional neural network process in order to make the deep-learning features complementary to radiomics features efficiently. 
Extensive experiments are conducted on a private dataset to verify the feasibility of our method. Experimental results suggest that combining the information of different features helps to train a superior model, which is a promising avenue for invasiveness prediction. 


{\small
	\bibliographystyle{unsrt}
	\bibliography{refs}
}

\end{document}